\documentclass[sigconf]{acmart}
\renewcommand\footnotetextcopyrightpermission[1]{} 

\usepackage[normalem]{ulem}

\usepackage{tikz}
\usepackage{multirow} 
\usepackage{siunitx}
\usepackage{makecell}
\usepackage{caption}
\usepackage{subcaption}
\usepackage{geometry}

\AtBeginDocument{%
  }

\begin{document}
\setlength{\floatsep}{1pt plus 0.5pt minus 0.5pt}
\setlength{\textfloatsep}{1pt plus 0.5pt minus 0.5pt}
\setlength{\intextsep}{1pt plus 0.5pt minus 0.5pt}
\setlength{\belowcaptionskip}{0.5pt}
\setlength{\belowcaptionskip}{0.5pt}

\title{An Efficient General-Purpose Optical Accelerator for Neural Networks}

\author{Sijie Fei}
\affiliation{%
  \institution{Technical University of Munich}
  \city{Munich}
  \country{Germany}}
\email{sijie.fei@tum.de}

\author{Amro Eldebiky}
\affiliation{%
  \institution{Technical University of Munich}
  \city{Munich}
  \country{Germany}}
\email{amro.eldebiky@tum.de}

\author{Grace Li Zhang}
\affiliation{%
  \institution{Technical University of Darmstadt}
  \city{Darmstadt}
  \country{Germany}}
  \email{grace.zhang@tu-darmstadt.de}

\author{Bing Li}
\affiliation{%
 \institution{University of Siegen}
 \city{Siegen}
 \country{Germany}}
 \email{Bing.Li@uni-siegen.de}

\author{Ulf Schlichtmann}
\affiliation{%
  \institution{Technical University of Munich}
  \city{Munich}
  \country{Germany}}
  \email{ulf.schlichtmann@tum.de}

\begin{abstract}
  General-purpose optical accelerators (GOAs) have emerged as a promising platform to accelerate deep neural networks (DNNs) due to their low latency and energy consumption. Such an accelerator is usually composed of a given number of interleaving Mach-Zehnder-Interferometers (MZIs). This interleaving architecture, however, has a low efficiency when accelerating neural networks of various sizes due to the mismatch between weight matrices and the GOA architecture. In this work, a hybrid GOA architecture is proposed to enhance the mapping efficiency of neural networks onto the GOA. In this architecture, independent MZI modules are connected with microring resonators (MRRs), so that they can be combined to process large neural networks efficiently. Each of these modules implements a unitary matrix with inputs adjusted by tunable coefficients. The parameters of the proposed architecture are searched using genetic algorithm. To enhance the accuracy of neural networks, selected weight matrices are expanded to multiple unitary matrices applying singular value decomposition (SVD). The kernels in neural networks are also adjusted to use up the on-chip computational resources. Experimental results show that with a given number of MZIs, the mapping efficiency of neural networks on the proposed architecture can be enhanced by 21.87\%, 21.20\%, 24.69\%, and 25.52\% for VGG16 and Resnet18 on datasets Cifar10 and Cifar100, respectively. The energy consumption and computation latency can also be reduced by over 67\% and 21\%, respectively.
\end{abstract}

\maketitle
\pagestyle{plain}

\section{Introduction}
Deep neural networks (DNNs) 
have achieved great breakthroughs and gained popularity in various applications. In order to tackle complex problems, the depth of neural networks has increased, which leads to a large number of multiply-accumulate (MAC) operations. Accordingly, efficient hardware platforms are highly demanded to accelerate such operations.  
Recently, analog in-memory-computing (IMC) platforms leveraging emerging technologies such as resistive RAM (RRAM) \cite{9116244,chi2016prime,9073995,10137106,8714954}, optical components \cite{onn,10323877,9256423} and Ferroelectric FET (FeFET) \cite{9474234} have been introduced. 
As a promising candidate, general-purpose optical accelerators (GOAs)  have emerged  to enhance the computational and energy efficiency of executing MAC operations in DNNs. In GOAs, weights are mapped onto Mach-Zehnder-Interferometer (MZI) arrays. Light signals representing input information propagate through MZI arrays to perform matrix-vector multiplications with a very high speed. The light signals carrying the computation results at the output of an MZI array can reach up to 100GHz, and the energy efficiency of such accelerators can be \SI{e+5} better than conventional GPUs \cite{onn}.

Despite high computational and energy efficiency, existing GOAs suffer from inefficient mapping. Such GOAs are usually composed of interleaving MZI arrays \cite{onn,tree,fft}, where thermal-optic phase shifters in MZIs are tuned to represent different weight matrices. Since the sizes of MZI arrays in a GOA are fixed after manufacturing, there is usually a mismatch between the dimensions of weight matrices and the GOA architecture. 
For example, 
when mapping small weight matrices onto a GOA with a large dimension, 
the MZIs in the GOA are not fully utilized.

Another challenge of existing GOAs is the large area cost. 
The matrices that the GOAs can represent are unitary. 
To represent an arbitrary weight matrix, the weight matrix should be decomposed using SVD \cite{onn} into two unitary matrices and one diagonal matrix before mapping. The GOAs can then realize the decomposed 
matrices directly.
For an $M\times N$ weight matrix, the number of MZIs needed in a GOA is around $\frac{M^2+N^2}{2}$. To reduce the large area overhead, new GOA architectures have been introduced. \cite{tree} replaces one unitary matrix with directional couplers of a tree architecture, reducing the number of MZIs to $\frac{N^2}{2}$. \cite{fft} proposes a Fast Fourier Transform (FFT) architecture for efficient computation and uses pruning techniques for further area reduction. 
In \cite{adept}, instead of using  MZIs, the architecture of GOAs adopts random combinations of directional couplers (DCs) and phase shifters (PSs) determined by reinforcement learning. 
However, MZIs and optical components are still organized in an interleaving manner and the inefficiency in mapping is still an issue. 

To enhance the efficiency of GOAs further, \cite{OplixNet} even encodes the input data onto amplitude and phase simultaneously so that the number of inputs and thus the size of the GOA array can be significantly reduced.

In this paper, we propose a hybrid architecture of  GOAs, with which 
the mapping  and area efficiency of neural networks can be improved significantly. The contributions of this paper are as follows:
\begin{itemize}
\item 
 A hybrid GOA architecture is proposed
to enhance the mapping efficiency of DNNs onto 
GOAs. In this architecture, independent MZI modules are connected
with microring resonators (MRR), so that they can be combined to
process large DNNs efficiently. 
\item 

 To reduce the area overhead, 
 each MZI module implements a unitary matrix with inputs adjusted by tunable coefficients. To enhance the utilization of the GOA architecture, 
a hardware-aware training is proposed, where the kernels in neural networks are derived from the SVD decomposition and the coefficients tuning the inputs of the MZI modules are adjusted to approximate the original weight matrices. Selective neural network expansion and weight matrix recovery are also deployed to maintain the accuracy of the neural networks.

\item 
To balance mapping efficiency, area efficiency, power consumption, and cost incurred by electrical/optical (E/O) conversion, the  parameters of the GOA architecture such as the size of MZI modules and their numbers are searched using Genetic Algorithm (GA) under different area limitations.

\end{itemize}

The structure of the remaining paper is as follows. Section \ref{preliminaries} introduces the preliminaries of MZIs and MRRs and the limitation of current mapping methods. Section \ref{proposed} proposes the new GOA architecture with the searched parameters and the framework to maintain inference accuracy. Experimental results are reported in Section \ref{Experimental Results}. Section \ref{conclusion} draws the conclusion.

\section{Preliminaries}
\label{preliminaries}
In this section,  
the functions of fundamental elements in the GOAs, e.g., MZI and MRR, will be introduced. The limitations in constructing GOAs with such elements will also be discussed. 
\subsection{ GOAs with MZIs and their limitations}
\begin{figure}[!t]
\centering
\captionsetup{font=bf}
\begin{tikzpicture}
\node[anchor=south west,inner sep=0] (image) at (0,0) {\includegraphics{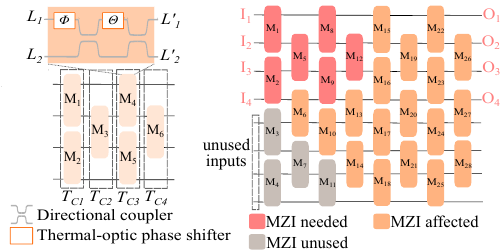}};
\begin{scope}[x={(image.south east)},y={(image.north west)}]
  \node[anchor=north west, fill=white] at (0.2,0.01) {(a)};
  \node[anchor=north east, fill=white] at (0.78,0.01) {(b)};
\end{scope}
\end{tikzpicture}
\captionsetup{skip=3pt}
\captionsetup{font=small}
\caption{(a) MZI structure and the interleaving MZI array for a 4$\times$4 unitary matrix. (b) Mapping a 4$\times$4 unitary matrix onto an 8$\times$8 interleaving MZI array.}
\label{mzi}
\end{figure}

An MZI is an optical component composed of two directional couplers and two thermal-optic phase shifters as shown in the upper part of Fig.~\ref{mzi}(a). The phase shifters are tuned by applying electrical currents. 
An MZI can implement a $2 \times 2$ matrix.  Two signals \( L_1 \) and \( L_2 \) are inputs.
Along the optical paths, they are altered by the phase shifters and interfere with each other. The transformed signals \( L_1' \) and \( L_2' \) are outputs from the two ends on the right in Fig.~\ref{mzi}(a). The transformation is explained with the following equation, where $\textbf{T}$ is called the transformation matrix and the superscript \( c \) indicates the complex domain.
\begin{equation*}
\begin{bmatrix}
 L_1'^c \\
 L_2'^c 
\end{bmatrix}
=
\begin{bmatrix}
 \frac{1}{\sqrt{2}}& \frac{i}{\sqrt{2}}\\
 \frac{i}{\sqrt{2}}& \frac{1}{\sqrt{2}}
\end{bmatrix}
\begin{bmatrix}
 e^{i\theta}& 0\\
 0& 1
\end{bmatrix}
\begin{bmatrix}
 \frac{1}{\sqrt{2}}& \frac{i}{\sqrt{2}}\\
 \frac{i}{\sqrt{2}}& \frac{1}{\sqrt{2}}
\end{bmatrix}
\begin{bmatrix}
 e^{i\phi}& 0\\
 0& 1
\end{bmatrix}
\begin{bmatrix}
 L_1^c \\
 L_2^c 
\end{bmatrix}
\label{equation1}
\end{equation*}

\begin{equation}
=ie^{\frac{i\theta}{2}}
\begin{bmatrix}
 e^{i\phi}\sin{\frac{\theta}{2}}& \cos{\frac{\theta}{2}}\\
 e^{i\phi}\cos{\frac{\theta}{2}}& -\sin{\frac{\theta}{2}}
\end{bmatrix}
\begin{bmatrix}
 L_1^c \\
 L_2^c 
\end{bmatrix}
\label{equation1}=\textbf{T}
\begin{bmatrix}
 L_1^c \\
 L_2^c 
\end{bmatrix}.
\end{equation}

To represent an $M\times N$ matrix, MZIs are arranged in an interleaving manner. The lower part of Fig.~\ref{mzi} (a) demonstrates the MZI array to realize a $4\times4$ unitary matrix. The matrix can be represented by the multiplication of the transformation matrices in all the columns as follows:  

\begin{equation}
\textbf{T} = \textbf{T}_{C_4}\textbf{T}_{C_3}\textbf{T}_{C_2}\textbf{T}_{C_1}
\label{equation3}
\end{equation}
\begin{equation}
\textbf{T}_{C_1}=
\begin{bmatrix}
\textbf{T}_1&0 \\
 0&\textbf{T}_2
\end{bmatrix}, \textbf{T}_{C_2}=
\begin{bmatrix}
 1& 0&0\\
 0& \textbf{T}_3&0\\
 0&0&1
\end{bmatrix}\hspace{5pt} \cdots
\label{equation4}
\end{equation}
where \(\textbf{T}_{C_1}\), \(\textbf{T}_{C_2}\), \(\textbf{T}_{C_3}\), and \(\textbf{T}_{C_4}\) represent transformation matrices of the first, second, third, and forth columns in Fig.~\ref{mzi}(a). \(\textbf{T}_1\),\(\textbf{T}_2\), and \(\textbf{T}_3\) are the transformation matrices of MZIs with labels \(\textbf{M}_1\), \(\textbf{M}_2\), and \(\textbf{M}_3\), respectively.

When a small matrix is mapped onto a GOA with a large dimension, the interleaving arrangement of MZIs has a low mapping efficiency. Fig.\ref{mzi} (b) shows an example of mapping a $4\times 4$ unitary matrix onto an $8\times 8$ MZI array, where $I_1$, $I_2$, $I_3$, $I_4$ denote the inputs and $O_1$, $O_2$, $O_3$, $O_4$ denote the outputs. For a $4 \times 4$ unitary matrix, only 6 MZIs on the upper left side are needed. However, due to the interleaving structure, 18 more MZIs along the signal paths are affected. These paths block the signals from the unused inputs. Accordingly, a large part of the MZIs in this GOA is actually wasted in this computation.
Because signals going through the MZIs that are affected 
may affect the valid outputs $O_1$, $O_2$, $O_3$, $O_4$, they also
need to be programmed correspondingly, although they do not contribute to the computation mathematically. This costs more unnecessary power consumption in applying GOAs to accelerate neural networks \cite{old_ps,ps}.


\subsection{GOAs with MRRs}
\begin{figure}[!t]
\centering
\captionsetup{font=bf}
\begin{tikzpicture}
\node[anchor=south west,inner sep=0] (image) at (0,0) {\includegraphics{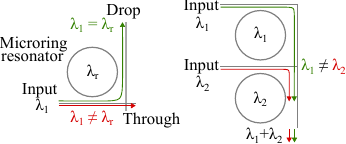}};
\begin{scope}[x={(image.south east)},y={(image.north west)}]
  \node[anchor=north west, fill=white] at (0.22,0.01) {(a)};
  \node[anchor=north east, fill=white] at (0.81,0.01) {(b)};
\end{scope}
\end{tikzpicture}
\captionsetup{skip=3pt}
\captionsetup{font=small}
\caption{MRR as (a) a switch and (b) an adder using WDM technique.}
\label{mrr}
\end{figure}

An MRR is a silicon-based ring, as depicted in Fig.~\ref{mrr} (a), characterized by a specific resonance wavelength \(\lambda_r\). When an input signal with wavelength \(\lambda_1\) propagates through the MRR, it can be directed to the drop waveguide if \(\lambda_1\) equals \(\lambda_r\); otherwise, the signal passes through the MRR following the original direction. By adjusting the MRR using a thermal tuner, its resonance wavelength can be modified. Therefore, an MRR can be utilized as an optical switch to control signal direction. 

Another function of MRRs is to add optical signals with different wavelengths using wavelength division multiplexing (WDM) technique. As shown in Fig.~\ref{mrr} (b), inputs with \(\lambda_1\) and \(\lambda_2\) are directed to the vertical fiber by the upper MRR and lower MRR, respectively. Because signals with wavelength \(\lambda_1\) are not affected by the lower MRR, the 
 two signals with different wavelengths reach the bottom of this column to be detected by a photodiode (PD), in fact implementing the summation of the two signals. In this structure, it is critical that the two MRRs work on different wavelengths to avoid the case that the signal from the top is diverted to the horizontal direction by the second MRR in Fig.~\ref{mrr} (b).

MRRs and WDM technique have also been used in realizing optical accelerators  \cite{mrr1,mrr2}. In such accelerators, inputs are modulated on light signals with different wavelengths. Weights are represented by the difference of transmission on the through port 
and the drop port 
of an MRR. Weighted signals processed by the MRRs can be accumulated using WDM and the summed outputs are detected using balanced photodiodes. However, because the number of wavelengths for WDM is limited \cite{mrr2}, the expressivity of accelerators based on MRR completely is limited.

\section{The Proposed Framework}
\label{proposed}
\label{section3}
To enhance the computational efficiency of MZI arrays, we propose a new GOA architecture, in which 
 independent small MZI modules are combined by MRRs to replace the previous large interleaving structure. MRRs are used 
 to accumulate the partial results computed by small MZI modules. To further mitigate the problem of large area cost, the weight matrices in neural network layers are approximated by unitary matrices, which can be directly implemented by MZI arrays, instead of requiring two unitary matrices as in previous methods \cite{onn,tree,fft}.
 For critical layers, the weight matrices can be selectively recovered back to the multiplication of unitary matrices to maintain the accuracy of neural networks.
 The parameters of the proposed GOA architecture are obtained by a search algorithm balancing the mapping efficiency of neural networks, area, power, and cost caused by electrical/optical conversions.
  Afterwards, neural networks are  adjusted to adapt to the GOA architecture to maximize the MZI utilization. The neural networks are further fine-tuned to counter the accuracy loss caused by the approximation using the unitary matrices implemented by the MZI modules.


\subsection{The proposed GOA architecture}

\begin{figure}[!t]
\captionsetup{font=bf}
\captionsetup{font=small}
\captionsetup{skip=3pt}
\centerline{\includegraphics{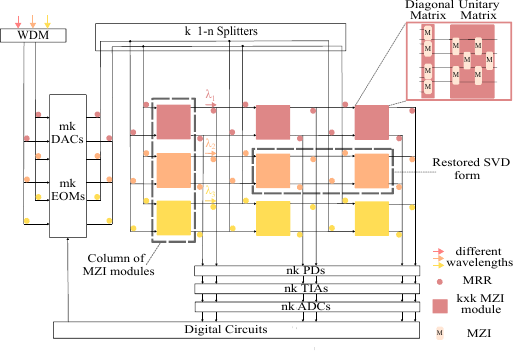}}
\caption{The proposed architecture, composing of small $k\times k$ MZI modules connected by MRRs and  peripheral devices, where \textit{m} and \textit{n} are the row number and column number of the MZI modules. }
\label{archi}
\end{figure}

Fig.~\ref{archi} illustrates the proposed GOA architecture  along with the peripheral optical devices.
$m$ and $n$ are the numbers of rows and columns of MZI modules in the GOA architecture, where different colors represent different wavelengths.
 The small $k\times k$ MZI modules are depicted as colored square blocks to
 process signals with specific wavelengths. Each MZI module is composed of a small MZI array whose inputs can be pre-modulated by a column of MZIs, equivalent to the multiplication with a diagonal matrix.
  MRRs are depicted as small circles, which
  can only redirect signals with corresponding resonant wavelengths.

As shown in Fig.~\ref{mzi}(a), a large MZI interleaving structure may lead to a  waste of resources. Therefore, in the proposed GOA architecture, we use the $k\times k$ small modules to construct large weight matrices.  For a general $M\times N$ weight matrix of a layer in a neural network, it is first partitioned into $k\times k$ submatrices. Each submatrix is implemented by a $k\times k$ module in Fig.~\ref{archi}. Since a $k\times k$ MZI interleaving structure can only realize a unitary matrix, a $k\times k$ weight submatrix \textit{$w$} normally needs to be decomposed using SVD \cite{onn} as
\begin{equation}\label{blocksvd}
w=U^{ \phantom{*} }_{svd}\Sigma^{ \phantom{*} }_{svd}V^{*}_{svd} 
\end{equation}
where $U^{ \phantom{*} }_{svd}$ and $V^{ \phantom{*} }_{svd}$ are unitary matrices and $\Sigma_{svd}$ is a diagonal matrix. 

\begin{figure}[!t]
\centering
\captionsetup{font=bf}
\begin{tikzpicture}
\node[anchor=south west,inner sep=0] (image) at (0,0) {\includegraphics{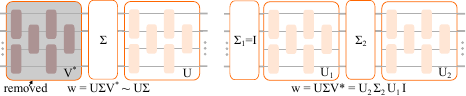}};
\begin{scope}[x={(image.south east)},y={(image.north west)}]
  \node[anchor=north west, fill=white] at (0.2,0.01) {(a)};
  \node[anchor=north east, fill=white] at (0.78,0.01) {(b)};
\end{scope}
\end{tikzpicture}
\captionsetup{skip=3pt}
\captionsetup{font=small}
\caption{(a) Matrix approximation with one unitary matrix. (b) Restoring the original matrix with two unitary matrices.}
\label{smallblock}
\end{figure}

To reduce computation cost, in the proposed GOA architecture the multiplication of the decomposed matrices $U^{ \phantom{*} }_{svd}\Sigma^{ \phantom{*} }_{svd}V^{*}_{svd}$ is approximated with $U\Sigma$, as shown in Fig.~\ref{smallblock}(a),
which is the structure of the basic modules in the proposed GOA architecture. Note that light signals enter such a module from the left and therefore the order of $U^{ \phantom{*} }_{svd}$, $\Sigma_{svd}$ and 
$V^{ \phantom{*} }_{svd}$ in Fig.~\ref{smallblock} is reversed compared with (\ref{blocksvd}).
Since the $\Sigma$ is a diagonal matrix, it can be implemented by a column of MZIs as shown at the upper right corner of Fig.~\ref{archi}. Compared with the SVD form, 
this approximation of a $k\times k$ weight matrix only needs one  $k\times k$ interleaving MZI structure and the number of MZIs is reduced by nearly a half.
For some $k\times k$ weight matrices, the approximation in Fig.~\ref{smallblock}(a) may lead to a relatively large error. To deal with this problem, we identify critical weight submatrices resulting from the partition of the original weight matrices of neural networks
and implement each of them with two modules, as shown in Fig.~\ref{smallblock}(b). 
In this case, the diagonal matrix inside the first module is set to an identity matrix, so that this combination of two modules in fact implements an SVD form in (\ref{blocksvd}).


 With the GOA architecture in Fig.~\ref{archi}, large weight matrices in neural networks can be implemented efficiently. 
  A column of modules in this architecture implements the sub weight matrices  
 in a row after partitioning the original weight matrix. 
 Their outputs are added, which is implemented by the MRRs at the outputs of each column in the GOA architecture. 
The MRRs direct the output signals to the bottom of the column, where the photodiodes detect the total energy of the accumulated light signals and convert them into electrical values representing the sum. 

In this accumulation process, the light signals from the outputs of the modules in a column are directed downwards. 
The signals at the top 
pass by other MRRs below in the same column.
The wavelengths of the lower MRRs
should be different from the wavelengths of the modules above. 
Otherwise, the light signals from above would be directed horizontally and cannot reach the photodiodes at the bottom. 
 Accordingly, it is required that all the modules in a column in the GOA architecture  
 process signals of different wavelengths, although the modules in the same row share the same wavelength.
  To meet this requirement, we use laser sources that generate different light signals to carry the input data to different modules in a column in the GOA, as shown on the left of Fig.~\ref{archi}. 
 



In the proposed GOA architecture, input signals are encoded with $mk$ digital-to-analog converters (DACs) and electrical-optical modulators (EOMs). To route the input signals to all the columns simultaneously as described above, 
each signal at the output of an EOM is split into $n$ signals by 1-$n$ splitters  and waveguides are created to connect them to the inputs of the corresponding modules.
At the bottom of the GOA, the outputs are detected and converted to digital signals by $nk$ PDs, transimpedance amplifiers (TIAs), and analog-to-digital converters (ADCs), so that they can be processed further in the digital domain.


\subsection{Mapping of neural networks and determining parameters of GOA}
\begin{figure}[!t]
\captionsetup{font=bf}
\captionsetup{font=small}
\captionsetup{skip=3pt}
\centerline{\includegraphics{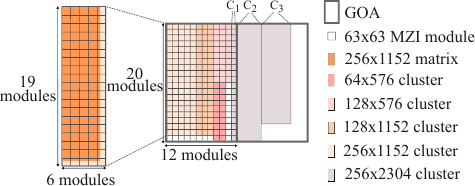}}
\caption{Mapping five weight matrices from VGG16 onto a GOA where \textit{m}=20, \textit{n}=12, \textit{k}=63. The GOA has 20 rows and 12 columns of 63$\times$63 MZI modules and the interconnections are omitted for simplicity.
On the left side, an 256$\times$1152 matrix is mapped onto 6$\times$19 MZI modules, called a 256$\times$1152 cluster. Mappings of clusters of different weight matrices are shown on the right side. The clusters are rotated by 90 degrees for mapping, to adapt to the input signals from the left side.}
\label{mapping}
\end{figure}

The architecture of the GOA is determined by the number of rows $m$  and the number of columns $n$ of the modules and the size of the modules $k\times k$. For a given area, or a given number of MZIs, $m$ and $n$ should be determined to maximize the mapping efficiency of neural networks. The mapping efficiency corresponds to how well the weight matrices are mapped onto the GOA, which affects the execution efficiency of neural networks.
To map a neural network onto a GOA, the weight matrices 
of convolutional layers are reshaped into $2D$ matrices. Specifically, $n$ filters of $h\times h$ kernels with depth $d$ are reshaped into a $n\times h^2d$ weight matrix. 
For example, the $7$th layer of VGG16 has 256 filters with $3\times 3$ kernels of depth 128. These kernels are reshaped into a $256\times1152$ matrix. 

To map a weight matrix of the size $n\times h^2d$, in total $\lceil \frac{n}{k} \rceil\times \lceil \frac{h^2d}{k} \rceil$  modules in the GOA are needed.
Fig.~\ref{mapping} illustrates how to map five weight matrices of VGG16 onto a GOA with $m$=20, $n$=12, $n$=63.
On the left side of Fig.~\ref{mapping}, $6\times 19$ modules are required to represent the reshaped $256\times1152$ weight matrix from VGG16, and the collective MZI modules for this weight matrix are called a $256\times1152$ cluster. Mapped clusters to represent different matrices are shown as the colored rectangles on the right side of Fig.~\ref{mapping}. Because the input signals come from the left side of the MZI modules, the clusters are rotated by $90$ degrees.

Because neural networks usually start with filters with small depths, the matrix clusters are usually of small dimensions. Therefore, for the initial mapping, it is feasible to arrange multiple matrix clusters together on a GOA to increase mapping efficiency. For example, in Fig.~\ref{mapping}, four mapping clusters exist on the GOA in the first mapping. 

According to the discussion above, the mapping efficiency and thus the execution efficiency of neural networks on the GOA are affected by the sizes of the filters in the neural networks and the parameters of the GOA. We use Genetic Algorithm (GA) to determine the parameters of the GOA with respect to the mapping efficiency of neural networks, area, power consumption, and E/O conversions. For simplicity, we use mapping cost to represent mapping efficiency, where a mapping cost of one neural network is defined as the total number of necessary mappings for this neural network on the given GOA. Lower mapping cost indicates higher mapping efficiency. 
The objective to optimize in GA is defined as follows:
\begin{equation}
\begin{split}
\textit{metric} &= \alpha\times \textit{mapping cost}+\beta\times \textit{area}\\
&+\gamma\times \textit{power}+\delta\times \textit{E/O conversions}
\end{split}
\label{metric}
\end{equation}
where $\alpha$, $\beta$, $\gamma$, and $\delta$ are the weighting factors of these objectives.

We use multiple neural networks to evaluate their average mapping cost in (\ref{metric}) with respect to a candidate of the GOA parameters in a search process. For a given candidate of the parameters of the GOA, i.e.,  a sample of the parameters $m$, $n$ and $k$ under a given number of MZIs,
the neural networks are mapped onto the GOA by combining as many filters onto the GOA as possible. 

The area and power are estimated by the area and power parameters of the MZIs, MRRs, DACs, EOMs, splitters, PDs, TIAs, and ADCs.  E/O conversions occur between layers of neural networks when weight clusters are segmented by the
GOA. For example, in Fig.~\ref{mapping}, the $256\times 2304$ cluster is segmented into three parts on the GOA and 
the computations are split into three submatrices, which introduces further computational latency. Therefore, E/O conversions are also minimized in the metric  (\ref{metric}).

\subsection{Neural network adjustment and hardware-aware training}

As illustrated in Fig.~\ref{mapping}, despite the compact mapping,  some MZIs remain underutilized due to the mismatch between the sizes of the weight matrices and the GOA. This can be addressed by increasing kernel depths and filter numbers of the layers of neural networks to take advantage of the unused resources to enhance the accuracy. In a neural network, the depth
 $d$ of one kernel is either equal to or proportional to -- due to the pooling layers -- the number of filters  in the previous layer. 
Therefore, when the kernel depth of one layer changes, the number of filters in the previous layer should change accordingly. 

\begin{figure}[!t]
\captionsetup{font=bf}
\captionsetup{font=small}
\captionsetup{skip=3pt}
\centerline{\includegraphics{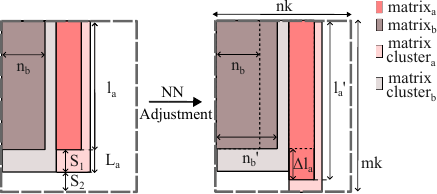}}
\caption{Adjusting the kernel depth of weight matrix $W_a$ and the filter number of the previous weight matrix $W_b$.}
\label{adjust}
\end{figure}

Fig.~\ref{adjust} shows an example of adjusting the weight matrices of two consecutive layers. The matrices are placed compactly on a GOA with length $mk$ and width $nk$, where the length is measured vertically and the width is measured horizontally. The matrix clusters to represent the matrices are depicted as colored rectangles.
For matrix $M_a$, it has length $l_a = h_a^2d_a$, where $h_a$ and $d_a$ are its kernel size and kernel depth, respectively.
The cluster size of this matrix has a length $L_a=\lceil\frac{l_a}{k}\rceil k$. 
The previous matrix $M_b$ has width $w_b=r\cdot d_a$,
with $r$ denoting the proportional ratio based on the pooling layer between the two layers. 

The available length increase $\Delta l_a$ of the weight matrix $M_a$ as shown in Fig.~\ref{adjust} is restricted as follows 
:
\begin{equation}
0\leq \Delta l_a \leq S_1+S_2
\label{length increase}
\end{equation}
\begin{equation}
S_1=L_a-l_a=\lceil\frac{l_a}{k}\rceil k-l_a, S_2 = mk-L_a = (m-\lceil\frac{l_a}{k}\rceil)k
\label{length increase2}
\end{equation}
where $S_1$ and $S_2$ are the unoccupied length in the matrix cluster and GOA, respectively. When the length of matrix $M_a$ increases by $\Delta l_a$, its kernel depth $d_a$ increases to $d_a'=\lfloor\frac{l_a+\Delta l_a}{h_a^2}\rfloor$ and the filter number of the previous matrix $M_b$ also increases to $rd_a'$.  
As a result, the length of the cluster of $M_a$ and the width of the cluster of $M_b$ are extended. Because changes in cluster widths can affect the mapping, the largest $\Delta l_a$ is selected while keeping the previous cluster width change as least as possible.

As illustrated in Fig.~\ref{smallblock}, 
a weight matrix in the SVD form is replaced by the multiplication of a diagonal matrix and a unitary matrix, which can lead to accuracy loss due to the approximation.
To enhance the accuracy of the neural network, a hardware-aware training is applied.
In the proposed architecture, a
 $k\times k$ matrix $w$ is approximated by $w'$ as follows:
\begin{equation}
\textit{$w'$} = \textit{$U$}\Sigma.
\label{transform}
\end{equation}
When training the neural network, we obtain $U$ as follows:
\begin{equation}
\textit{$U$} = \textit{$U_{svd}^{ \phantom{*} }$}\textit{$V_{svd}^{*}$},  \text{ where } \textit{$w$}=\textit{$U_{svd}^{ \phantom{*} }$}\Sigma_{svd}^{ \phantom{*} }\textit{$V^{*}_{svd}$} 
\label{uv}
\end{equation}
and 
\begin{equation*}
\Sigma=
\begin{pmatrix}
\sigma_{1,1} & \cdots & 0 \\
\vdots & \ddots & \vdots \\
0 & \cdots & \sigma_{k,k}
\end{pmatrix}.
\end{equation*}
To identify the values $\sigma_{1,1},...\sigma_{k,k}$ in $\Sigma$, we minimize the distance between the approximated matrix $w'=\textit{U}\Sigma$ and the original matrix $w$ as
\begin{equation}\label{mindiag}
\sigma_{k,k} = \operatorname*{argmin}_{\sigma}(\|\textit{$w_{k}$}-\sigma_{k,k}\textit{$U_k$}\|_2)
\end{equation}
where $w_k$ and $U_k$ are the $k$th rows of  the corresponding matrices, respectively.
The transformation (\ref{transform})--(\ref{mindiag}) is applied every $p$ epochs during the hardware-aware training to guarantee the training efficiency while keeping the weight matrix in the approximated form  (\ref{transform}) as close to the original matrix as possible.


Beyond the compensation and training above, we also identify weight matrices that are critical to the accuracy of the neural network and recover them back to the original multiplication form shown in (\ref{blocksvd}). This is possible in the proposed GOA architecture shown in Fig.~\ref{archi} because the modules are also connected horizontally. If a matrix needs to be recovered, it is implemented by two neighboring modules by configuring the MRRs between them to allow horizontal signal propagation instead of directing the signals downwards. In this way, the critical matrices are kept in their original form to maintain the accuracy.


The matrices are restored by columns to maintain the regular shape of clusters, namely a column of small modules is either restored together or stays unchanged.
To identify the matrices to recover, the columns are sorted according to the accumulated error within their MZI modules and a given number of columns with the largest error are selected to restore. 
If a column to be restored is located on the GOA edge and cannot be connected with a subsequent column, it is skipped and the column with the next largest error is selected. 
After columns are restored, hardware-aware training is conducted again to enhance the accuracy of the neural network.


\section{Experimental Results}
\label{Experimental-Results}
To evaluate the proposed framework, two different neural networks, VGG16 \cite{vgg}, and Resnet18 \cite{res} were tested on two datasets, Cifar10 and Cifar100 \cite{cifar100}. VGG16 and Resnet18 were trained with Nvidia Quadro RTX 6000 GPUs  from scratch for 250 epochs and 300 epochs, respectively.
The training epoch interval $p$ for matrix transformation as described in Section~\ref{section3} for the two neural networks were every 5 and 10 epochs, respectively. The GOA architecture was searched based on the estimated area and power parameters of optical devices listed in Table~\ref{device params}. The weighting factors in (\ref{metric}) of mapping cost, area, power, and E/O conversions were set to 1, 0.8, 0.8, and 0.2.

\begin{table}[!t]
\small
\captionsetup{font=bf}
\captionsetup{skip=3pt}
\captionsetup{font=small}
\caption{Optical Device Parameters}
\begin{center}
\begin{tabular}{|c|c|c|}
\hline
\textbf{Device}&\textbf{Parameter}&\textbf{Value}\\
\hline
\multirow{2}{*}{\makecell{PS}} & operation latency&  \SI{5.2}{\micro\second}\\
&tuning power, area&\SI{10}{\milli\watt}/$\pi$, \SI{100}{\micro\meter\squared}\cite{ps}\\
\hline
DC &area &\SI{130}{\micro\meter\squared}\cite{dc}\\

\hline
MRR & holding power, area & \SI{3.1}{\milli\watt}\cite{mrr-power}, \SI{7.07}{\micro\meter\squared}\cite{mrr-area}\\
\hline
Laser & power, area & \SI{37.5}{\milli\watt}, \num{1000} $\times$ \SI{6}{\micro\meter\squared}\cite{laser}\\
\hline
DAC & power, area & \SI{26}{\milli\watt}, \SI{0.06}{\milli\meter\squared}\cite{dac}\\
\hline
EOM & power, area & \SI{1}{\milli\watt}, \SI{1000}{\micro\meter\squared}\cite{eom}\\
\hline
\multirow{2}{*}{PD} & responsivity, voltage & \SI{26}{\milli\ampere}/W, \SI{2}{\volt}\\
 &length & \SI{40}{\micro\meter}\cite{pd}\\
\hline
TIA & power, area & \SI{3}{\milli\watt}, \SI{0.015}{\milli\meter\squared}\cite{tia}\\
\hline
ADC & power, area & \SI{29}{\milli\watt}, \SI{0.103}{\milli\meter\squared}\cite{adc}\\
 
\hline
\end{tabular}
\label{device params}
\end{center}
\end{table}

\begin{table*}[!t]
\small
\captionsetup{font=bf}
\captionsetup{font=small}
\captionsetup{skip=3pt}
\caption{Results of the proposed framework. MZI number constraint: 20000, \textit{m}, \textit{n}, \textit{k} = 6,3,44}

\begin{center}
\begin{tabular}{c c c c c c c c c c}
\toprule
\multirow{3}{*}{{\makecell{\textbf{Neural Networks}\\\textbf{Dataset}}}}&\multicolumn{3}{c}{\textbf{Performance Improvement}}& \multicolumn{1}{c!{\vrule width 0pt}}{}&\multicolumn{4}{c}{\textbf{Accuracy}}&\multirow{3}{*}{{\makecell{\textbf{Restored}\\\textbf{Cols}}}}\\

\cline{2-4}\cline{6-9}
& \multirow{2}{*}{{\makecell{\textbf{Mapping}\\\textbf{Reduction}}}}&\multirow{2}{*}{{\makecell{\textbf{Energy}\\\textbf{Reduction}}}}& \multirow{2}{*}{{\makecell{\textbf{Latency}\\\textbf{Reduction}}}}& \multicolumn{1}{c!{\vrule width 0pt}}{}& \multirow{2}{*}{\textbf{Baseline}}
& \multirow{2}{*}{{\makecell{\textbf{This Work}\\ \textbf{w/o adjustment}}}}& \multirow{2}{*}{{\makecell{\textbf{This Work}\\ \textbf{w/o restoration}}}} & \multirow{2}{*}{{\makecell{\textbf{This Work}}}} &\\
& &&& \multicolumn{1}{c!{\vrule width 0pt}}{}& \multirow{2}{*}& &\\

\midrule
VGG16-Cifar10& 21.87\%&67.96\%&21.85\%&&93.55\%&92.57\%&92.67\%&93.57\%& 1\%\\

VGG16-Cifar100& 21.20\%&67.71\%&21.19\%&&70.16\%&67.12\%&68.35\%&70.67\%& 2\%\\

Resnet18-Cifar10 & 24.69\%&69.13\%&24.61\%&&94.93\%&94.91\%&94.94\%&95.22\%& 1\%\\

Resnet18-Cifar100 & 25.52\%&69.47\%&25.45\%&&75.79\%&75.94\%&76.44\%&76.44\%& 0\%\\

\bottomrule
\end{tabular}
\label{main table}
\end{center}
\end{table*}

Table \ref{main table} shows the energy and latency improvement, as well as the improved accuracy, compared with the original MZI interleaving architecture, when the target number of MZIs was set to  20000. The searched $m$, $n$, and $k$ were 6, 3, and 44, respectively. The first column lists the tested neural networks and datasets. The second column shows the reduction of mappings for each neural network. Because  neural networks have different architectures, e.g., numbers of filters and feature maps, they have different mapping efficiencies on the same GOA. Furthermore, restorations of MZI module columns to maintain accuracy on different datasets also affects the mapping efficiency.
Compared with the interleaving architecture,
mapping cost for VGG16 and Resnet18 for datasets Cifar10 and Cifar100 were reduced by 21.87\%, 21.20\%, 24.69\%, and 25.52\%, respectively. 
With reduced mapping, energy and latency are also reduced, as shown in the third and the forth columns in Table \ref{main table}. For all neural networks, the energy reductions are above 67\% and the latency reductions are above 21\%. 
The fifth column in Table \ref{main table} shows the baseline accuracy when the neural networks are trained directly. The sixth column shows the accuracy of neural networks after implementing the hardware-aware training without adjustments to neural network structures or column restoration. For dataset Cifar100, compared with the baselines, VGG16 is affected by the absence of one unitary matrix, leading to accuracy degradation while the accuracy of the other test cases is well maintained. 
The seventh column shows the accuracy after applying structural adjustments in neural networks but without restoring any column. Compared with the sixth column, the accuracy is enhanced around 1\% for VGG16-Cifar100, owing to the increased expressivity by increased kernel depths and filter numbers. The eighth column shows the accuracy using the complete framework with neural network adjustment and column restoration, which is better than the baseline accuracy consistently. The percentage of the columns that are restored is shown in 
the last column, which is negligible in all these test cases.



\begin{figure}[!t]
\centering
\captionsetup{font=bf}
\captionsetup{font=small}
\begin{tikzpicture}
\node[anchor=south west,inner sep=0] (image) at (0,0) {\includegraphics{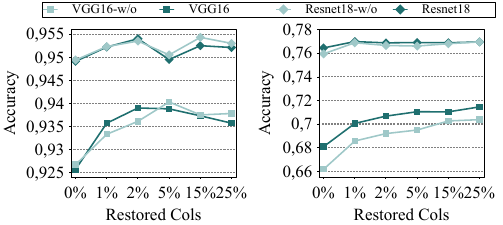}};
\begin{scope}[x={(image.south east)},y={(image.north west)}]
  \node[anchor=north west, fill=white] at (0.27,0.01) {(a)};
  \node[anchor=north east, fill=white] at (0.85,0.01) {(b)};
\end{scope}
\end{tikzpicture}
\captionsetup{skip=3pt}
\caption{Accuracy with respect to the number of restored columns without and with neural network adjustments for datasets. (a) VGG16 and Resnet18 on Cifar10 and (b) VGG16 and Resnet18 on Cifar100.}
\label{accuracy-restore-2}
\end{figure}

To demonstrate the effect of hardware-aware training and column restoration, 
Fig.~\ref{accuracy-restore-2} shows the accuracy improvement with respect to these two techniques. In Fig.~\ref{accuracy-restore-2}(a), the results are obtained with VGG16 and Resnet18 on Cifar10. The x-axis shows the number of columns that are restored. The y-axis shows the accuracy of the neural networks. The different dashed curves compare the neural networks with and without structural adjustments.
According to Fig.~\ref{accuracy-restore-2}(a), the accuracy improvement after adjusting the neural network architectures is not obvious for Cifar10. However, only by restoring 1\% to 2\% of columns, accuracy can be improved by around 1\%. 
In Fig.~\ref{accuracy-restore-2}(b) the results on Cifar100 are shown. In these cases,
network architecture adjustment can achieve a better gain in accuracy, 
e.g., 2\% accuracy improvement for VGG16. In addition,
restoration of columns also has a better effect on Cifar100.
With more restored columns, the accuracy can achieve a higher improvement but the increase trend is gradually flatted. This is because the latter restored columns already have negligible errors. 

\begin{figure}[!t]
\centering
\captionsetup{font=bf}
\captionsetup{font=small}
\begin{tikzpicture}
\node[anchor=south west,inner sep=0] (image) at (0,0) {\includegraphics{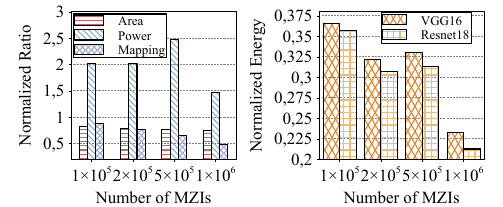}};
\begin{scope}[x={(image.south east)},y={(image.north west)}]
  \node[anchor=north west, fill=white] at (0.28,0.01) {(a)};
  \node[anchor=north east, fill=white] at (0.84,0.01) {(b)};
\end{scope}
\end{tikzpicture}
\captionsetup{skip=3pt}
\caption{(a) Normalized area, power and mapping over SVD accelerator and (b) Normalized energy consumption for running a neural network when 10000, 20000, 50000, 100000 MZIs are used.}
\label{svd}
\end{figure}

To evaluate the improved performance of the proposed architecture, it is compared with the interleaving SVD accelerator with a structure shown in Fig.~\ref{mzi} and the Adept 16x16-a4 accelerator in \cite{adept}. 
Fig.~\ref{svd} shows the comparison of this work with the SVD interleaving accelerator.  To ensure a fair comparison, the inputs and outputs of the SVD accelerator are kept the same as the proposed architecture. The overall area cost, power for optical computations, and the average number of mapping of VGG16 and Resnet18 are shown for comparison. Fig.~\ref{svd} (a) illustrates the normalized values of this work over an SVD accelerator when 10000, 20000, 50000, 100000 MZIs are used. The restored columns for each neural network are the same as listed in Table \ref{main table}. 
Although introducing MRRs and more lasers, the proposed work has a better area efficiency because the number of MZIs is reduced efficiently. The overall area costs can be reduced by around 18\% to 25\%. Compared with the interleaving structure of the SVD accelerator, the independent MZI modules in this work can achieve mapping reductions from 13\% to 52\%. 
Due to the extra holding power of MRRs and increased power of more lasers, the total power also increases. This power increase, however, can be offset by the substantial reduction in mapping costs, leading to overall energy savings as shown in Fig.~\ref{svd} (b), where the 
normalized energy consumption over the SVD accelerator for running VGG16 and Resnet18 covering optical computations and weight matrices mapping is shown. 
Despite the increased power of extra MRRs and lasers, the reduced mapping effort still leads to significant energy reduction, resulting in overall energy saving from 64\% to 77\%.


\begin{figure}[!t]
\centering
\captionsetup{font=bf}
\captionsetup{font=small}
\begin{tikzpicture}
\node[anchor=south west,inner sep=0] (image) at (0,0) {\includegraphics{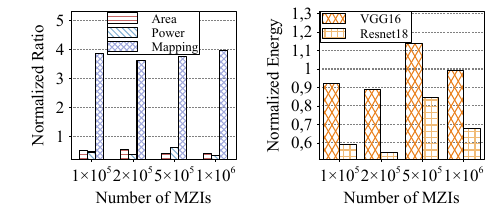}};
\begin{scope}[x={(image.south east)},y={(image.north west)}]
  \node[anchor=north west, fill=white] at (0.27,0.01) {(a)};
  \node[anchor=north east, fill=white] at (0.85,0.01) {(b)};
\end{scope}
\end{tikzpicture}
\captionsetup{skip=3pt}
\caption{(a) Normalized area, power and mappings over the Adept accelerator and (b) Normalized energy consumption for processing 1000 pictures when 10000, 20000, 50000, 100000 MZIs are used.}
\label{adept}
\end{figure}

In Fig.~\ref{adept}, the proposed architecture is evaluated against the Adept 16$\times$16-a4 accelerator \cite{adept}. The small MZI modules in the proposed architecture are replaced by the 16$\times$16 Photonic Tensor Cores (PTC), which comprises of 174 crossings, 71 DCs and 10 unitary blocks \cite{adept}. The total number of MZIs remains the same, and the area for one PTC is estimated by the footprint ratio in \cite{adept}. The dimensions of the PTCs are fixed to 16$\times$16, and the row number $m$ and column number $n$ are determined by traversing all potential candidates.

Fig.~\ref{adept} (a) presents the normalized area, power, and average mapping of the proposed accelerator over the Adept accelerator. The proposed architecture demonstrates area efficiency improvement from 46\% to 60\% and power efficiency improvement from 38\% to 66\%, which is enabled by further modules sizes beyond the Adept accelerator returned by the searching process - an indicator that more efficient dimensions may also exist for the Adept accelerator.
Because the Adept PTC has a lower area cost than the proposed orthogonal form, under the same MZI area, it provides more small MZI modules. Therefore,the Adept accelerator has a lower mapping effort compared with the proposed work.
Nevertheless, when processing multiple images, the proposed architecture can achieve lower energy consumption. Fig.~\ref{adept} (b) shows the normalized energy consumption of the proposed architecture over the Adept accelerators when processing 1000 pictures. Because of the low operation frequency and high operation power of thermal-optic phase-shifters, it is more efficient to process multiple
images in batch. In this scenario, the significant operation power reduction of this work leads to a lower energy consumption in most cases.


\begin{figure}[!t]
\centering
\captionsetup{font=bf}
\captionsetup{font=small}
\begin{tikzpicture}
\node[anchor=south west,inner sep=0] (image) at (0,0) {\includegraphics{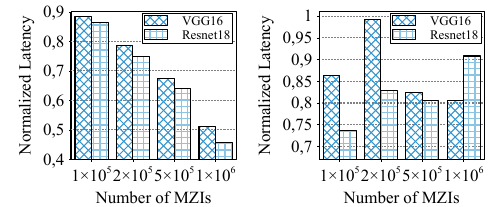}};
\begin{scope}[x={(image.south east)},y={(image.north west)}]
  \node[anchor=north west, fill=white] at (0.27,0.01) {(a)};
  \node[anchor=north east, fill=white] at (0.85,0.01) {(b)};
\end{scope}
\end{tikzpicture}
\captionsetup{skip=3pt}
\caption{Normalized computation latency (a) over SVD accelerator and (b) over the Adept accelerator.}
\label{latency}
\end{figure}

Fig.~\ref{latency} (a) and (b) compare the computation latency of the proposed architecture with the SVD accelerator and the Adept accelerator, respectively. The proposed work has a lower latency than the SVD accelerator because of the reduced mapping effort. Compared with the Adept accelerator, the latency in this work is reduced with fewer E/O conversions. Frequent E/O conversions result in segmented weight matrices and should be avoided in parameter search as much as possible. However, the structure of the Adept accelerators is predetermined without incorporating E/O conversions, thus resulting in a higher latency. 

\section{Conclusion}
\label{conclusion}
A hybrid architecture was proposed to determine the structure of GOA and 
reduce mapping effort of neural networks. 
Traditional interleaving MZI arrays are split into small MZI modules and connected with MRRs. This architecture allows multiple small matrices to be mapped onto a GOA together without mutual interferce to improve mapping efficiency. To further minimize  area cost, one unitary matrix is reduced in each small MZI module. Afterwards, the architecture parameters are searched using GA, balancing the mapping efficiency, area, power, and cost incurred by E/O conversions. To improve MZI utilization, the kernels in the neural networks are adjusted to fit into the GOA. Then a hardware-training is applied. Errors caused by the absence of one unitary matrix are compensated by expanding critical weight matrices back to the SVD format.
Experimental results show that the mapping effort can be reduced by 21.87\%, 21.20\%,
24.69\%, and 25.52\% for VGG16 and Resnet18 on datasets Cifar10 and
Cifar100, respectively. 
Reductions on energy consumption and computation latency can reach above 66\% and 21\%, respectively.

\bibliographystyle{unsrt}
\bibliography{sample-base}

\begin{thebibliography}{10}

\bibitem{9116244}
Ying Zhu, Grace~Li Zhang, Tianchen Wang, Bing Li, Yiyu Shi, Tsung-Yi Ho, and
  Ulf Schlichtmann.
\newblock Statistical training for neuromorphic computing using memristor-based
  crossbars considering process variations and noise.
\newblock In {\em Design, Automation \& Test in Europe Conference \& Exhibition
  (DATE)}, pages 1590--1593, 2020.

\bibitem{chi2016prime}
Ping Chi, Shuangchen Li, Cong Xu, Tao Zhang, Jishen Zhao, Yongpan Liu, Yu~Wang,
  and Yuan Xie.
\newblock Prime: A novel processing-in-memory architecture for neural network
  computation in {R}e{RAM}-based main memory.
\newblock In {\em International Symposium on Computer Architecture (ISCA)},
  2016.

\bibitem{9073995}
Shuhang Zhang, Grace~Li Zhang, Bing Li, Hai~Helen Li, and Ulf Schlichtmann.
\newblock Lifetime enhancement for rram-based computing-in-memory engine
  considering aging and thermal effects.
\newblock In {\em IEEE International Conference on Artificial Intelligence
  Circuits and Systems (AICAS)}, pages 11--15, 2020.

\bibitem{10137106}
Amro Eldebiky, Grace~Li Zhang, Georg Boecherer, Bing Li, and Ulf Schlichtmann.
\newblock Correct{N}et: Robustness enhancement of analog in-memory computing
  for neural networks by error suppression and compensation.
\newblock In {\em Design, Automation \& Test in Europe Conference \& Exhibition
  (DATE)}, 2023.

\bibitem{8714954}
Shuhang Zhang, Grace~Li Zhang, Bing Li, Hai~Helen Li, and Ulf Schlichtmann.
\newblock Aging-aware lifetime enhancement for memristor-based neuromorphic
  computing.
\newblock In {\em Design, Automation \& Test in Europe Conference \& Exhibition
  (DATE)}, pages 1751--1756, 2019.

\bibitem{onn}
Yichen Shen, Nicholas~C Harris, Skirlo, et~al.
\newblock Deep learning with coherent nanophotonic circuits.
\newblock {\em Nature photonics}, 11(7):441--446, 2017.

\bibitem{10323877}
Amro Eldebiky, Bing Li, and Grace~Li Zhang.
\newblock Nearuni: Near-unitary training for efficient optical neural networks.
\newblock In {\em International Conference on Computer Aided Design (ICCAD)},
  pages 1--8, 2023.

\bibitem{9256423}
Ying Zhu, Grace~Li Zhang, Bing Li, Xunzhao Yin, Cheng Zhuo, Huaxi Gu, Tsung-Yi
  Ho, and Ulf Schlichtmann.
\newblock Countering variations and thermal effects for accurate optical neural
  networks.
\newblock In {\em International Conference On Computer Aided Design (ICCAD)},
  pages 1--7, 2020.

\bibitem{9474234}
Yu~Qian, Zhenhao Fan, Haoran Wang, Chao Li, Mohsen Imani, Kai Ni, Grace~Li
  Zhang, Bing Li, Ulf Schlichtmann, Cheng Zhuo, and Xunzhao Yin.
\newblock Energy-aware designs of ferroelectric ternary content addressable
  memory.
\newblock In {\em Design, Automation \& Test in Europe Conference \& Exhibition
  (DATE)}, pages 1090--1095, 2021.

\bibitem{tree}
Zheng Zhao, Derong Liu, Meng Li, et~al.
\newblock Hardware-software co-design of slimmed optical neural networks.
\newblock In {\em Asia and South Pacific Design Automation Conference
  (ASPDAC)}, 2019.

\bibitem{fft}
Jiaqi Gu, Zheng Zhao, Chenghao Feng, et~al.
\newblock Towards area-efficient optical neural networks: {An} {FFT}-based
  architecture.
\newblock In {\em Asia and South Pacific Design Automation Conference
  (ASPDAC)}, 2020.

\bibitem{adept}
Jiaqi Gu, Hanqing Zhu, Chenghao Feng, et~al.
\newblock Adept: {Automatic} differentiable design of photonic tensor cores.
\newblock In {\em Design Automation Conference (DAC)}, 2022.

\bibitem{OplixNet}
Ruidi Qiu, Amro Eldebiky, Li~Zhang, et~al.
\newblock {OplixNet}: {Towards} area-efficient optical split-complex networks
  with real-to-complex data assignment and knowledge distillation.
\newblock In {\em Design, Automation and Test in Europe (DATE)}, 2024.

\bibitem{old_ps}
Nicholas~C Harris, Yangjin Ma, Jacob Mower, et~al.
\newblock Efficient, compact and low loss thermo-optic phase shifter in
  silicon.
\newblock {\em Optics express}, 22(9):10487--10493, 2014.

\bibitem{ps}
Jorge Parra, Juan Hurtado, Amadeu Griol, et~al.
\newblock Ultra-low loss hybrid {ITO/Si} thermo-optic phase shifter with
  optimized power consumption.
\newblock {\em Optics express}, 28(7):9393--9404, 2020.

\bibitem{mrr1}
Johannes Feldmann, Nathan Youngblood, C~David Wright, et~al.
\newblock All-optical spiking neurosynaptic networks with self-learning
  capabilities.
\newblock {\em Nature}, 569(7755):208--214, 2019.

\bibitem{mrr2}
Viraj Bangari, Bicky~A Marquez, Heidi Miller, et~al.
\newblock Digital electronics and analog photonics for convolutional neural
  networks ({DEAP-CNNs}).
\newblock {\em IEEE Journal of Selected Topics in Quantum Electronics},
  26(1):1--13, 2019.

\bibitem{vgg}
Karen Simonyan and Andrew Zisserman.
\newblock Very deep convolutional networks for large-scale image recognition.
\newblock {\em ArXiv}, 2014.

\bibitem{res}
Kaiming He, Xiangyu Zhang, Shaoqing Ren, et~al.
\newblock Deep residual learning for image recognition.
\newblock In {\em Conference on Computer Vision and Pattern Recognition
  (CVPR)}, 2016.

\bibitem{cifar100}
Alex Krizhevsky and Geoffrey Hinton.
\newblock Learning multiple layers of features from tiny images.
\newblock 2009.

\bibitem{dc}
Luhua Xu, Yun Wang, Amar Kumar, et~al.
\newblock Compact high-performance adiabatic 3-{dB} coupler enabled by
  subwavelength grating slot in the silicon-on-insulator platform.
\newblock {\em Optics express}, 26(23):29873--29885, 2018.

\bibitem{mrr-power}
Po~Dong, Wei Qian, Hong Liang, et~al.
\newblock Thermally tunable silicon racetrack resonators with ultralow tuning
  power.
\newblock {\em Optics express}, 18(19):20298--20304, 2010.

\bibitem{mrr-area}
Qianfan Xu, David Fattal, and Raymond~G Beausoleil.
\newblock Silicon microring resonators with 1.5-$\mu$m radius.
\newblock {\em Optics express}, 16(6):4309--4315, 2008.

\bibitem{laser}
A~Descos, C~Jany, D~Bordel, et~al.
\newblock Heterogeneously integrated {III-V/Si} distributed bragg reflector
  laser with adiabatic coupling.
\newblock In {\em European Conference and Exhibition on Optical Communication
  (ECOC)}, 2013.

\bibitem{dac}
Behnam Sedighi, Mahdi Khafaji, and J~Christoph Scheytt.
\newblock 8-bit 5{GS/s} {D/A} converter for multi-gigabit wireless
  transceivers.
\newblock In {\em European Microwave Integrated Circuit Conference (EuMIC)},
  2011.

\bibitem{eom}
Po~Dong, Shirong Liao, Dazeng Feng, et~al.
\newblock Low {Vpp}, ultralow-energy, compact, high-speed silicon electro-optic
  modulator.
\newblock {\em Optics express}, 17(25):22484--22490, 2009.

\bibitem{pd}
Zhen Sheng, Liu Liu, Joost Brouckaert, et~al.
\newblock {InGaAs} {PIN} photodetectors integrated on silicon-on-insulator
  waveguides.
\newblock {\em Optics express}, 18(2):1756--1761, 2010.

\bibitem{tia}
Michal Rakowski, Yoojin Ban, Peter De~Heyn, et~al.
\newblock Hybrid 14nm {FinFET}-silicon photonics technology for low-power
  tb/s/mm$^2$ optical {I/O}.
\newblock In {\em Symposium on VLSI Technology}, 2018.

\bibitem{adc}
Mingqiang Guo, Jiaji Mao, Sai-Weng Sin, et~al.
\newblock A 5 {GS/s} 29 mw interleaved {SAR} {ADC} with 48.5 db {SNDR} using
  digital-mixing background timing-skew calibration for direct sampling
  applications.
\newblock {\em IEEE Access}, 8:138944--138954, 2020.

\end{thebibliography}

\end{document}